\documentclass[10pt,twocolumn,letterpaper]{article}

\usepackage{cvpr}              
\definecolor{cvprblue}{rgb}{0.21,0.49,0.74}
\usepackage[pagebackref,breaklinks,colorlinks,allcolors=cvprblue]{hyperref}

\usepackage{cuted}
\usepackage{caption} 

\usepackage[utf8]{inputenc} 
\usepackage[T1]{fontenc}    
\usepackage{hyperref}       
\usepackage{url}            
\usepackage{booktabs}       
\usepackage{amsfonts}       
\usepackage{nicefrac}       
\usepackage{microtype}      
\usepackage{xcolor}         
\usepackage{colortbl}
\definecolor{customcolor}{HTML}{D3D3D3}
\usepackage{amsmath} 
\usepackage{multirow}
\usepackage{microtype}
\usepackage{graphicx}

\usepackage{booktabs} 
\usepackage[ruled,vlined,linesnumbered]{algorithm2e}
\usepackage{colortbl}
\usepackage{wrapfig}

\def\paperID{12202} 
\def\confName{CVPR}
\def\confYear{2026}

\title{Towards Open Environments and Instructions: General Vision-Language Navigation via Fast-Slow Interactive Reasoning}

\author{Yang Li\textsuperscript{1}, Aming Wu\textsuperscript{2}, Zihao Zhang\textsuperscript{1}, Yahong Han\textsuperscript{1}\thanks{Corresponding author.}\\
\textsuperscript{1}School of Artificial Intelligence, College of Intelligence and Computing, Tianjin University, China\\
\textsuperscript{2}School of Computer Science and Information Engineering, Hefei University of Technology, China\\
{\tt\small liyang1389@tju.edu.cn, zhangzihao2490@tju.edu.cn, amwu@hfut.edu.cn,  yahong@tju.edu.cn}
}

\begin{document}

\maketitle

\begin{strip}
    \centering
    \includegraphics[width=0.9\textwidth]{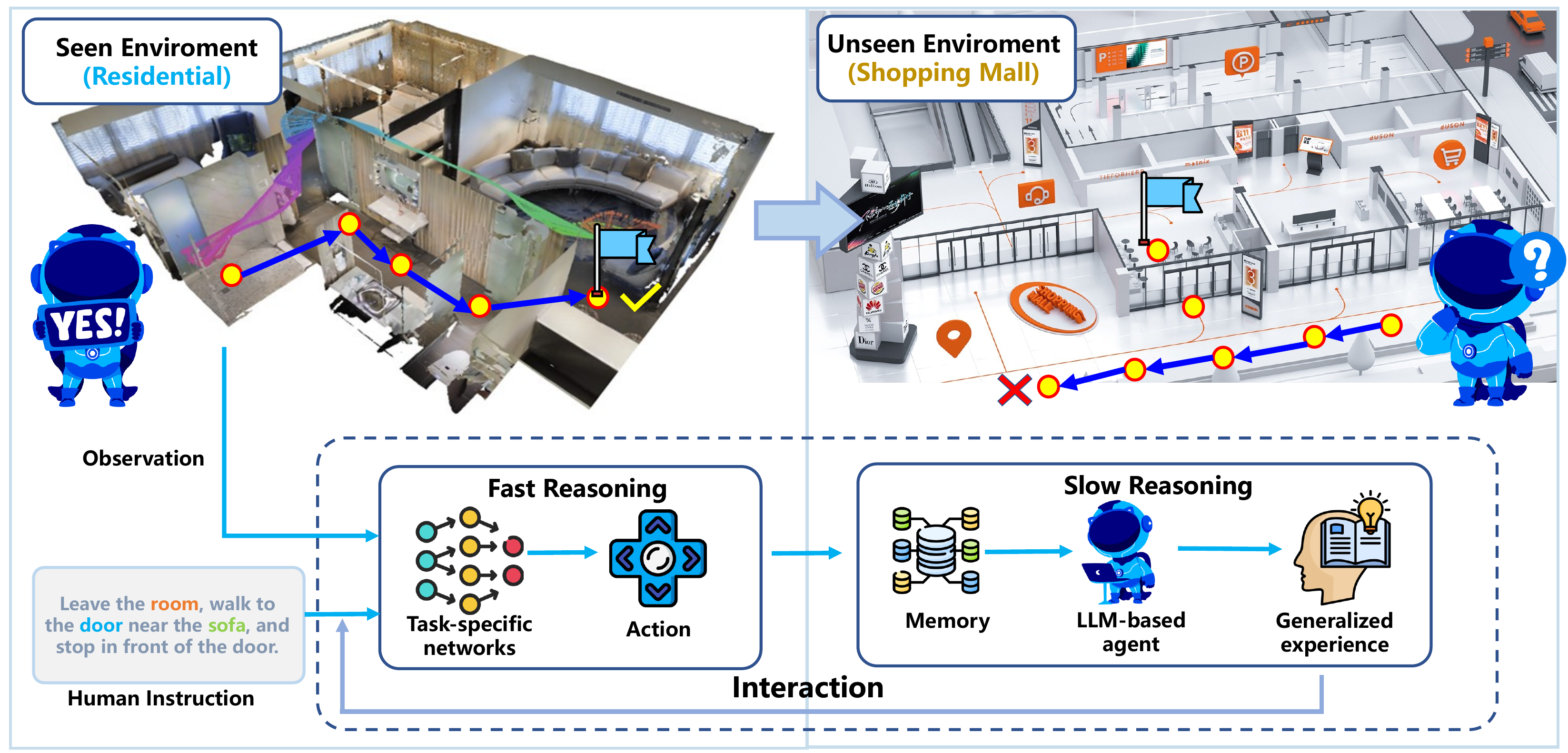}
    \captionof{figure}{In the GSA-VLN task, the training set focuses on residential environments, while the test set includes non-residential scenes such as shopping malls, offices, and cinemas. It also incorporates basic, scene, and user-style instructions. The core goal is to evaluate the agent’s scene generalization ability through diverse building types and instruction variations. To address this open-world navigation challenge, we propose the interactive Slow4Fast framework: “fast reasoning” is driven by a policy network that outputs actions from real-time input and stores memories; “slow reasoning” processes memories, extract generalized experiences, and reinforce the policy network.}
    \label{fig:teaser}
    \vspace{10pt} 
\end{strip}

\begin{abstract}
\vspace{-15pt}

Vision-Language Navigation (VLN) aims to enable agents to navigate to a target location based on language instructions. Traditional VLN often follows a close-set assumption, i.e., training and test data share the same style of the input images and instructions. 
However, the real world is open and filled with various unseen environments, posing enormous difficulties for close-set methods.
To this end, we focus on the General Scene Adaptation (GSA-VLN) task, aiming to learn generalized navigation ability by introducing diverse environments and inconsistent instructions.

Towards this task, when facing unseen environments and instructions, the challenge mainly lies in how to enable the agent to dynamically produce generalized strategies during the navigation process. 
Recent research indicates that by means of fast and slow cognition systems, human beings could generate stable policies, which strengthen their adaptation for open world. 
Inspired by this idea, we propose the slow4fast-VLN, establishing a dynamic interactive fast-slow reasoning framework.
The fast-reasoning module, an end-to-end strategy network, outputs actions via real-time input. It accumulates execution records in a history repository to build memory.
The slow-reasoning module analyze the memories generated by the fast-reasoning module. 
Through deep reflection, it extracts experiences that enhance the generalization ability of decision-making. These experiences are structurally stored and used to continuously optimize the fast-reasoning module. Unlike traditional methods that treat fast-slow reasoning as independent mechanisms, our framework enables fast-slow interaction. By leveraging the experiences from slow reasoning, it continually improves the accuracy and generalization ability of fast decisions. This interaction allows the system to continuously adapt and efficiently execute navigation tasks when facing unseen scenarios. Extensive experiments demonstrate the superiorities of our method.
\end{abstract}    

\section{Introduction}
Vision-Language Navigation (VLN)~\citep{anderson2018vision} is a fundamental task in embodied AI, enabling robots to operate in real-world environments.
Traditional VLN approaches like those in~\citep{qi2020reverie, thomason2020visioncvdn}, often follow a closed-set assumption, where both the training and test data share similar styles of environments and instructions.
However, it fails to capture the complexity of real-world scenarios, where environments are dynamic and instructions vary widely in style and context. In practice, agents must navigate previously unseen environments, posing a significant challenge for closed-set methods, which struggle to adapt to new settings with differing environmental and instructional contexts.
To this end, we focus on the recently proposed GSA-VLN (General Scene Adaptation for VLN) task~\citep{GSA-VLN}, aiming to learn generalized navigation ability by introducing diverse environments and inconsistent intructions.
Towards this task, when facing unseen environments and instructions, the challenge mainly lies in how to enable the agent to dynamically produce generalized strategies during the navigation process.

Research~\citep{Aming1, Aming2, wang2026vggdrive, Causal-NVD} shows that when agents transition from familiar testing environments to complex real-world settings, it can lead to spurious reasoning pathways, similar to hallucinations. This makes it difficult for agents to recognize their limitations or uncertainties~\citep{chen2023hallucination,Aming3,Aming4}. The root cause lies in the lack of explicit modeling of fast and slow cognitive processes, reminiscent of human \textit{System 1} and \textit{System 2} thinking~\citep{yao2024tree, thinkingfast}. Recent research~\citep{Fastandslow1, Fastandslow3} indicates that through dual cognition systems, agents can generate stable policies that enhance their adaptation to the open world.

However, existing methods that use dual systems \cite{FastandSlow4,FastandSlow6} often design fast and slow system as two independent, parallel systems that handle different types of tasks respectively. This fragmented structure lacks information interaction, which limits its application in the GSA-VLN task: 
Although slow reasoning can solve complex scenarios, the experience it gains cannot be consolidated into the strategies of the fast reasoning network. The result is that fast reasoning always remains at its original level, and when facing similar scenarios, it still needs to repeatedly invoke slow reasoning, lacking performance improvement that evolves over time. In open worlds and unseen scenarios, generalized experience cannot be compressed into low-latency intuitive response patterns. This causes the agent to still perform like a “novice driver” in out-of-distribution scenarios, with weakened generalization and adaptation capabilities. 




To this end, we propose the slow4fast-VLN framework, establishing a dynamic interactive fast-slow reasoning framework. The fast-reasoning module is an end-to-end strategy network that directly outputs actions based on real-time observations and instructions. It accumulates execution records in the history repository, building memory to face familiar environments and instructions.
The slow-reasoning module refines history repository into structured knowledge by analyzing key successes and failures. Using a large language model (LLM), it reflects on and extracts generalizable experiences related to scenarios, storing them in the experience library and used to continuously optimize the fast-reasoning module. 
This interaction allows the system to continuously adapt and efficiently execute navigation tasks when facing unseen scenarios. 
Additionally, GR-DUET~\citep{GSA-VLN} focuses on scene adaptation from a visual perspective but overlooks adapting to diverse instruction styles. We address this by implementing instruction style transformation through Chain-of-Thought prompt engineering to capture consistent speaking styles within a fixed environment.


\vspace{-5pt}

\section{Fast-Slow Interactive Reasoning}
\subsection{Task Definition of GSA-VLN}
Traditional VLN requires an agent to follow a language instruction $I$ to navigate from a start viewpoint to the target viewpoint.  
At timestep $t$, the agent receives a panoramic observation $O_{t}$ containing $K$ single-view observations $O_{t,k}$, i.e., $O_{t}=\{O_{t,k}\}_{k=1}^{K}$.
There are $N$ navigable views 
among $K$ views. The navigable views and a stop token $[stop]$ form the action space, from which the agent chooses one as the action prediction $a_{t}$.
The GSA-VLN task integrates multiple datasets, covering 150 scenes and 20 building types, while clearly distinguishing between in-distribution (ID) and out-of-distribution (OOD) scenes to test the agent's ability to adapt to unfamiliar scene types. At the same time, unlike the simple instructions in regular VLN tasks, the instructions in GSA-VLN are more diverse, simulating real user language habits. They are divided into three categories: basic, scene-specific and personalized user instructions, covering various language styles.

\subsection{Overall Framework}

Research~\citep{GSA-VLN} has shown that existing navigation methods perform poorly in OOD environments. The essential reason mainly lies in the instability of the reasoning process of the agent in complex scenarios and OOD scenarios. Therefore, the core issue is how to effectively improve the robustness of the reasoning process. In general, when facing familiar scenarios, humans rely on fast-thinking for quick decision-making, while for unfamiliar scenarios, we engage in slow-thinking to analyze and internalize the experience as a foundation for future fast thinking, which strengthens the adaptation for unknown world. 

\textbf{Definition 1} \textit{(System 1 and System 2)} \textit{System 1 and System 2 are two distinct reasoning systems proposed by Daniel Kahneman in his book Thinking, Fast and Slow~\citep{thinkingfast}.}

\textit{\textcolor{blue}{System 1 (Fast Thinking):} Refers to an unconscious, automatic thinking process that is fast, intuitive, and effortless. It is responsible for automatic responses and basic cognitive operations in daily activities but is susceptible to heuristic biases and errors.}

\textit{
\textcolor{blue}{System 2 (Slow Thinking):} Refers to a conscious, effortful thinking process that is slow, demanding, logical. It is responsible for complex calculations, reasoning, and decision-making processes.}



\begin{figure*}[t]
    \centering
    \includegraphics[width=0.9\linewidth]{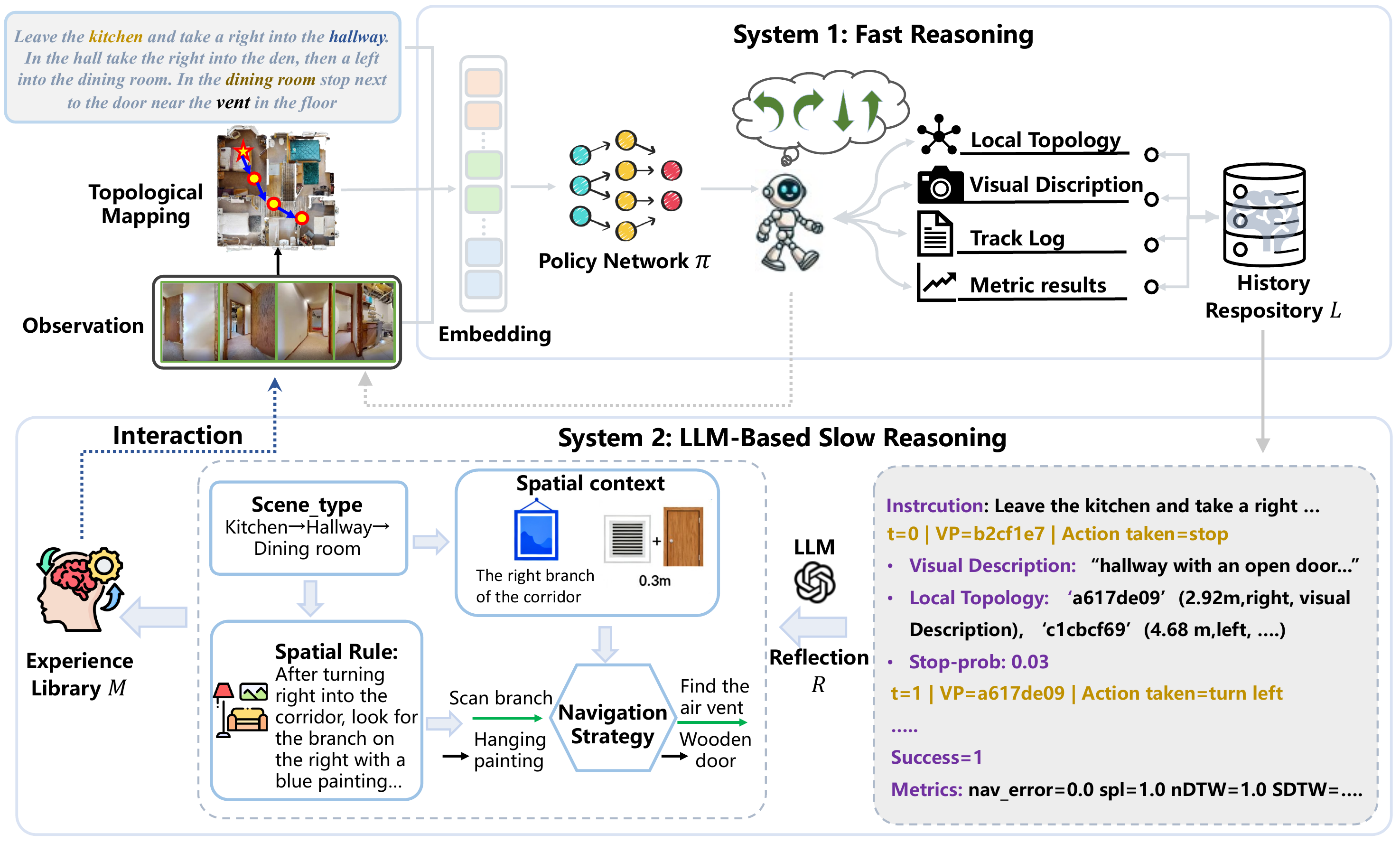}

    \caption{\textbf{Overview of our method.} The policy network processes real-time input, executes actions, and stores historical memory. The slow-reasoning network reflects on these memories to generate generalized experiences, which are then stored. These experiences guide the fast-reasoning network, providing strategic insights when faced with complex scenarios.
    }
    \label{overview}
     
\end{figure*}

Inspired by this, we propose fast-slow interactive reasoning to improve the generalization of navigation (see Fig. \ref{overview}). The fast-reasoning network processes real-time input, executes actions, and stores historical memory. The slow-reasoning network reflects on these memories to generate generalized experiences. These experiences guide the fast-reasoning network, providing strategic insights when faced with complex scenarios.
Formally, the framework can be expressed as an iterative process:
\begin{equation}
\mathcal{F} = \langle \pi, R, M, A \rangle,
\end{equation}
where \( \pi \) represents the policy network for executing fast reasoning , \( R \) is the reflection function, \( M \) is the experience extraction and storage module, and \( A \) is used to empower the fast-reasoning network with generalizable experience. The process of each episode \( k \) is:
\begin{equation}
\begin{split}
L_k = \pi(I_k, Env), \quad R_k = R(L_k), \\
\quad \mathcal{E}_k = M(R_k), \quad \pi_{k+1} = A(\pi_k, \mathcal{E}_k),
\end{split}
\end{equation}
where \( I_k \) is the instruction of the \( k \)-th episode, \( Env \) is the environment, \( L_k \) is the generated history memory, and \( \mathcal{E}_k \) is the extracted experience set.

\subsection{Fast Reasoning}
Concretely, the fast-reasoning system is a policy network $\pi$. Here, we adopt DUET~\citep{chen2022think} architecture. The input consists of instructions, current environmental observation (including panoramic images, GPS location, and neighbor node information), and historical navigation data. A topology mapping module dynamically constructs and updates a map with visited, navigable, and current nodes based on historical data. 
The global action planning module performs dual-scale encoding: the coarse-scale encoder provides global navigation scores, and the fine-scale encoder generates local actions. The dynamic fusion module then computes fusion weights to select the highest-scoring node as the next action.
In addition, for each node, we process its visual features using the Llama3.2-vision~\citep{llama} to generate a textual description of the viewpoint, with each node in the topology map having its corresponding description. During navigation, a history trajectory is produced and stored as memory. A navigation episode is represented by the time step sequence $\mathcal{T} = \{t_1, t_2, \dots, t_N\}$, where $N$ is the total number of steps. The historical data $\mathcal{L}(t_j)$ is defined as:
\begin{equation}
\mathcal{L}(t_j) = \left[ t_j, j_{\text{seq}}, V_j, \mathcal{T}_{\text{local}}, I, A_j^s, F_v(j), \mathcal{U}_{\text{step}} \right]^\top .
\label{logdata}
\end{equation}
For each time step $t_j$, the historical data $\mathcal{L}(t_j)$ includes the timestamp $t_j$, step sequence $j_{\text{seq}}$, viewpoint $V_j$, local topology $\mathcal{T}_{\text{local}}$ (with neighboring viewpoints, azimuth, and distance), navigation instruction $I$, selected action $A_j^s$, visual description $F_v(j)$, and step metrics $\mathcal{U}_{\text{step}}$ (such as stop probability and trajectory effectiveness). This data tracks the agent’s progress throughout the episode.

While the fast reasoning module can handle navigation in most familiar scenarios, research~\citep{wu2023textit, ge2024openagi} shows that when agents transition from familiar testing environments to OOD scenarios, they may generate spurious reasoning pathways. This makes it difficult for agents to recognize their own limitations or uncertainties. The fundamental reason lies in the lack of explicit modeling of slow cognitive processes. Recent research~\citep{Fastandslow1, Fastandslow2} indicates that by modeling slow cognitive systems, agents can develop stable strategies, thereby enhancing their adaptability to the open world. Therefore, we will introduce the slow-reasoning process.

\subsection{Slow Reasoning}
The slow-reasoning framework is used to convert fast-reasoning memories into structured, generalized experiences stored in the Experience Library. These experiences are integrated into the policy network for dynamic strategy adjustments, enabling continuous performance optimization.


\textbf{Experience Structure Design.} Structured experience $\mathcal{E}$ is defined as a vector:
\begin{equation}
\mathcal{E} = \left[ S_t, C_s, R_s, T_n, \eta_s, f \right]^\top,
\label{experience-template}
\end{equation}
where $S_t$ represents the scene type, $C_s$ denotes spatial context, $R_s$ indicates spatial rules, $T_n$ is the navigation strategy, $\eta_s$ is the historical success rate, and $f$ is the frequency of occurrence. These components collectively characterize key information and knowledge in the navigation process.



\textbf{Chain-of-Thought Prompt for Reflection.} 
This study designs a structured chain-of-thought (CoT) reflection prompt template \( \mathcal{P} \) to guide LLM in extracting valuable experience from navigation data. The template simulates human reasoning through a logical, step-by-step process: 
\begin{equation}
\mathcal{P}(\mathcal{X}) = \mathcal{P}_{\text{intro}} + \mathcal{P}_{\text{ctx}}(\mathcal{X}) + \mathcal{P}_{\text{tasks}} + \mathcal{P}_{\text{output}} .
\end{equation}
The template includes key components: the role definition and task guidance module $\mathcal{P}_{\text{intro}}$, which sets the LLM’s role and core task; the context-filling module $\mathcal{P}_{\text{ctx}}(\mathcal{X})$, which provides necessary navigation data for analysis; the task decomposition module $\mathcal{P}_{\text{tasks}}$, which breaks down experience extraction into subtasks like scene recognition and navigation strategy analysis; and the output format constraint module $\mathcal{P}_{\text{output}}$, which ensures structured results. This prompt enhances the LLM's ability to analyze and extract valuable experience from navigation data. The detailed prompt template can be seen in Supplementary Material.
The experience generation process is modeled as the LLM mapping function \( \mathcal{F}_{\text{LLM}} \), where the input is the reflection prompt \( \mathcal{P}(\mathcal{X}) \) and the output is the experience \( \mathcal{E} \):
\begin{equation}
\mathcal{E} = \mathcal{F}_{\text{LLM}}(\mathcal{P}(\mathcal{X})) .
\end{equation}
Existing methods~\citep{Fastandslow1, Fastandslow2} often design fast and slow systems as independent parallel structures for different tasks. While slow reasoning handles complex scenarios, its experience cannot integrate into fast strategies, leaving fast reasoning stagnant at the initial level.
When facing similar scenarios, the slow reasoning process needs to be repeatedly invoked, which undermines the real-time navigation requirements.
In fact, slow thinking should not be considered a one-time solution to complex problems. Its true value lies in produce generalized strategies that can enhance the fast-thinking system. The goal is to empower the agent to solve most problems efficiently, relying primarily on fast thinking while maintaining the flexibility to adapt to novel and unseen environments. Therefore, we establish a dynamic interactive fast-slow reasoning framework.





\subsection{Interaction between Fast-Slow Reasoning}

The fast reasoning module processes real-time navigation inputs and stores historical memories. Then, the slow reasoning module reflects on and summarizes them, extracting generalizable experiences to deal with OOD scenarios. The key here is how to achieve fast-slow interaction? How to use slow reasoning to empower fast reasoning networks. To this end, we adopt the following solution: the slow reasoning module retrieves experiences related to the current scenario from the experience library and encodes them into specific vectors. Next, the visual features of the fast reasoning network $\pi$ are fused with this vector through attention. And the experience-enhanced navigation decisions are output, which is the interaction between fast and slow reasoning.

\textbf{Experience Retrieval and Encoding}. The experience library $M = \{ E_1, E_2, \dots, E_K \}$ is a finite set of experiences with a capacity of $K$. 
At the current timestep of navigation decision-making, let the current context be $\mathcal{X}_{\text{cur}}$ (including information such as the current scene, spatial location). First, extract the retrieval key features $\mathcal{K} = [S_t^{\text{cur}}, C_s^{\text{cur}}, T_n^{\text{cur}}]$ from $\mathcal{X}_{\text{cur}}$, then calculate the feature similarity $\text{sim}(\mathcal{K}, \mathcal{E}_i)$ between $\mathcal{K}$ and all $\mathcal{E}_i$ in the $M$. 
Let the retrieval threshold be $\tau_{\text{retrieve}}$, and select the experiences with a similarity greater than or equal to $\tau_{\text{retrieve}}$, sorted in descending order of similarity. The most relevant $M$ experiences are selected to form the experience set $M_{\text{sel}}$.
In order to align the experiences with the feature of the fast reasoning network $\pi$, an experience encoder $G_{\text{enc}}$ is designed to convert each selected experience $E_{\text{sel},k}$ into a vector representation $F_e(k) \in \mathbb{R}^d$, where $d$ is the experience embedding dimension. For discrete features $S_t, C_s, T_n$, they are converted into vectors through an embedding layer, with embedding dimension $d/3$. A linear layer and activation function are applied to obtain the final experience embedding. For the $M$ experiences in $M_{\text{sel}}$, after encoding, the set of experience features $F_e = \{F_e(1), \dots, F_e(M)\}$ is obtained.

\textbf{Experience Fusion}. Let the original visual feature of the fast reasoning network $\pi$ be $F_v \in \mathbb{R}^{B \times L \times D}$ (where $B$ is the batch size, $L$ is the number of views, and $D$ is the dimension of visual features). The core of the interaction between fast and slow reasoning is to fuse $F_v$ and $F_e$ through an attention mechanism.
First, we expand $F_e$ to a dimension consistent with the batch size of $F_v$, resulting in $F_e^{\text{exp}} \in \mathbb{R}^{B \times M \times d}$. Then, we calculate the attention weights between visual features and experience features using a multi-head attention layer:
\begin{equation}
F_{\text{att}}, \omega = \text{MultiHeadAttn}(Q=F_v, K=F_e^{\text{exp}}, V=F_e^{\text{exp}}),
\end{equation}
where $\omega$ is the attention weight, and $F_{\text{att}} \in \mathbb{R}^{B \times L \times d}$ is the experience feature after attention weighting.
Next, we perform a concatenation operation on $F_v$ and $F_{\text{att}}$ along the feature dimension, and map the result back to the feature dimension of fast reasoning network $\pi$ via a linear layer:
\begin{equation}
F_{\text{fused}} = \sigma\left( W_{\text{fusion}} \cdot [F_v; F_{\text{att}}] + b_{\text{fusion}} \right),
\end{equation}
where $W_{\text{fusion}} \in \mathbb{R}^{D \times (D + d)}$ and $b_{\text{fusion}} \in \mathbb{R}^D$ are the parameters of the fusion layer.
Finally, we replace the original visual feature of the fast reasoning network $\pi$ with $F_{\text{fused}}$, and input it into the model for forward computation to obtain the experience-empowered navigation decision output $Y_{\text{enhanced}}$:
\begin{equation}
Y_{\text{enhanced}} = \pi(F_{\text{fused}}, I),
\end{equation}
where $I$ is the navigation instruction and $Y_{\text{enhanced}}$ including the action probability distribution and navigation confidence.

\subsection{Instruction Style Conversion}
Using a LLM model as the foundation, instruction style conversion is implemented through Chain-of-Thought prompt engineering. The specific process is as follows: when the system receives User or Scene style instructions, it first constructs a prompt text containing style conversion requirements. This prompt is then input into the LLM model, which automatically identifies and converts stylistic features in the instructions based on its language understanding capabilities, while preserving the core navigation semantics of the instruction unchanged. Finally, it outputs the converted Basic-style instruction. The system also computes a confidence score for the conversion; if the confidence exceeds a preset threshold, the converted instruction is used, otherwise the original instruction is retained. This entire process occurs in real-time during navigation training without requiring additional pre-training phases. Through this approach, dynamic conversion from Scene and User style instructions to Basic style is achieved, providing the navigation model with uniformly formatted instruction inputs. The detailed prompt template content can be seen in Supplementary Material.

\begin{table*}
  \centering
  \vspace{-8pt}
  \caption{Comparison of different adaptation methods in GSA-R2R with basic instructions.}
  \label{tab:basic}
  \resizebox{\textwidth}{!}{
  \begin{tabular}{l|ccccc|ccccc}
    \toprule
    \multicolumn{1}{c|}{\multirow{2}{*}[-0.5ex]{\textbf{Methods}}} & \multicolumn{5}{c|}{\textbf{Test-R-Basic}} & \multicolumn{5}{c}{\textbf{Test-N-Basic}}  \\
    \cline{2-11}
    & TL & NE$\downarrow$  & SR$\uparrow$ & SPL$\uparrow$  & nDTW$\uparrow$ & TL & NE$\downarrow$  & SR$\uparrow$ & SPL$\uparrow$  & nDTW$\uparrow$  \\
    \hline
    \multicolumn{11}{c}{\emph{Baseline}} \\
    DUET~\citep{chen2022think}            & 13.1 & 4.2 & 57.7 & 47.0 & 55.6 & 14.8 & 5.3 & 48.1 & 37.3 & 45.9  \\
    \hline
    \multicolumn{11}{c}{\emph{Optimization-Based Methods}} \\   

    \quad +MLM~\citep{devlin2018bert}      & 13.1 \scriptsize{$\pm$0.1} & 4.1 \scriptsize{$\pm$0.1} & 57.9 \scriptsize{$\pm$0.2} & 47.3 \scriptsize{$\pm$0.1} & 55.9 \scriptsize{$\pm$0.2} & 13.1 \scriptsize{$\pm$0.2} & 5.3 \scriptsize{$\pm$0.1} & 48.3 \scriptsize{$\pm$0.5} & 38.8 \scriptsize{$\pm$0.5} & 48.4 \scriptsize{$\pm$0.3} \\
        
    \quad +MRC~\citep{lu2019vilbert}      & 13.1 \scriptsize{$\pm$0.1} & 4.2 \scriptsize{$\pm$0.1} & 57.7 \scriptsize{$\pm$0.1} & 47.0 \scriptsize{$\pm$0.1} & 55.6 \scriptsize{$\pm$0.1} & 14.7 \scriptsize{$\pm$0.1} & 5.3 \scriptsize{$\pm$0.1} & 48.1 \scriptsize{$\pm$0.1} & 37.3 \scriptsize{$\pm$0.1} & 45.9 \scriptsize{$\pm$0.1} \\

    \quad +BT~\citep{wang2020active}       & 8.0 \scriptsize{$\pm$0.1} & 3.8 \scriptsize{$\pm$0.1} & 61.3 \scriptsize{$\pm$0.6} & 57.7 \scriptsize{$\pm$0.3}& 70.1 \scriptsize{$\pm$0.5}& 7.9 \scriptsize{$\pm$0.0} & 5.2 \scriptsize{$\pm$0.1} & 49.5 \scriptsize{$\pm$0.8} & 46.0 \scriptsize{$\pm$0.8} & 59.4 \scriptsize{$\pm$0.9}  \\
    
    \quad +TENT~\citep{wang2021tent} & 14.6 \scriptsize{$\pm$0.0} & 4.2 \scriptsize{$\pm$0.0} & 57.2 \scriptsize{$\pm$0.4} & 44.2 \scriptsize{$\pm$0.4} & 52.9 \scriptsize{$\pm$0.1} & 16.2 \scriptsize{$\pm$0.1} & 5.4 \scriptsize{$\pm$0.1} & 46.5 \scriptsize{$\pm$0.4} & 33.7 \scriptsize{$\pm$0.2} & 42.6 \scriptsize{$\pm$0.3} \\
    
    \quad +SAR~\citep{niu2023towards} & 13.8 \scriptsize{$\pm$0.8} & 4.0 \scriptsize{$\pm$0.1} & 57.6 \scriptsize{$\pm$0.2} & 44.6 \scriptsize{$\pm$0.2} & 53.0 \scriptsize{$\pm$0.2} & 16.5 \scriptsize{$\pm$0.0} & 5.4 \scriptsize{$\pm$0.0} & 44.6 \scriptsize{$\pm$1.5} & 31.5 \scriptsize{$\pm$1.6} & 40.6 \scriptsize{$\pm$1.3} \\
    
    \hline
    \multicolumn{11}{c}{\emph{Memory-Based Methods}} \\   
    
    TourHAMT~\citep{krantz2023iterative}   & 11.6 \scriptsize{$\pm$0.1} & 7.4 \scriptsize{$\pm$0.1} & 14.9 \scriptsize{$\pm$0.1} & 12.2 \scriptsize{$\pm$0.1} & 34.7 \scriptsize{$\pm$0.1} & 9.4 \scriptsize{$\pm$0.1} & 7.7 \scriptsize{$\pm$0.1} & 11.0 \scriptsize{$\pm$0.2} & 8.6 \scriptsize{$\pm$0.2} & 32.2 \scriptsize{$\pm$0.1}  \\
    
    OVER-NAV~\citep{zhao2024over}        & 14.1 \scriptsize{$\pm$0.1} & 6.7 \scriptsize{$\pm$0.0} & 22.3 \scriptsize{$\pm$0.3} & 16.8 \scriptsize{$\pm$0.2} & 37.1 \scriptsize{$\pm$0.1} & 11.4 \scriptsize{$\pm$0.1} & 7.1 \scriptsize{$\pm$0.1} & 16.6 \scriptsize{$\pm$0.2} & 13.0 \scriptsize{$\pm$0.1} & 35.0 \scriptsize{$\pm$0.2} \\
    
    GR-DUET~\citep{GSA-VLN}  & 9.4 \scriptsize{$\pm$0.0} & 3.1 \scriptsize{$\pm$0.0} & 69.3 \scriptsize{$\pm$0.2} & 64.3 \scriptsize{$\pm$0.1} & 71.4 \scriptsize{$\pm$0.1} & 8.9 \scriptsize{$\pm$0.0} & 4.4 \scriptsize{$\pm$0.0} & 56.6 \scriptsize{$\pm$0.1} & 51.5 \scriptsize{$\pm$0.1} & 61.0 \scriptsize{$\pm$0.1} \\
    Ours  & 9.6 \scriptsize{$\pm$0.1} & \textbf{2.9} \scriptsize{$\pm$0.2} & \textbf{70.8} \scriptsize{$\pm$0.1} & \textbf{65.0} \scriptsize{$\pm$0.1} & \textbf{72.1} \scriptsize{$\pm$0.1} & 10.2 \scriptsize{$\pm$0.1} & \textbf{4.2} \scriptsize{$\pm$0.0} & \textbf{58.4} \scriptsize{$\pm$0.2} & \textbf{52.9} \scriptsize{$\pm$0.1} & \textbf{62.4} \scriptsize{$\pm$0.3} \\

    \bottomrule
  \end{tabular}
  }
\end{table*}

\begin{table*}
  \centering
  \caption{Comparison of different adaptation methods in GSA-R2R with User instructions.}
  \label{tab:user}
  \resizebox{\textwidth}{!}{ 
  \begin{tabular}{l|cc|cc|cc|cc|cc}
    \toprule
    \multicolumn{1}{c|}{\multirow{2}{*}[-0.5ex]{\textbf{Methods}}} & \multicolumn{2}{c|}{\textbf{Child}} & \multicolumn{2}{c|}{\textbf{Keith}} & \multicolumn{2}{c|}{\textbf{Moira}} & \multicolumn{2}{c|}{\textbf{Rachel}} & \multicolumn{2}{c}{\textbf{Sheldon}}\\
    \cline{2-11}
    & SR$\uparrow$ & SPL$\uparrow$  & SR$\uparrow$ & SPL$\uparrow$  & SR$\uparrow$ & SPL$\uparrow$ & SR$\uparrow$ & SPL$\uparrow$  & SR$\uparrow$ & SPL$\uparrow$  \\
    \hline
    \multicolumn{11}{c}{\emph{Baseline}} \\
    DUET       & 54.3 & 44.1 & 56.0 & 46.3 & 52.3 & 43.3 & 56.3 & 46.4 & 54.0 & 44.4 \\
    \hline
    \multicolumn{11}{c}{\emph{Optimization-Based Methods}} \\   
    \quad +MLM & 54.5 \scriptsize{$\pm$0.2} & 44.7 \scriptsize{$\pm$0.2} & 56.4 \scriptsize{$\pm$0.3} & 46.8 \scriptsize{$\pm$0.3} & 53.8 \scriptsize{$\pm$0.3} & 43.6 \scriptsize{$\pm$0.4} & 56.8 \scriptsize{$\pm$0.5} & 46.6 \scriptsize{$\pm$0.6} & 54.5 \scriptsize{$\pm$0.4} & 44.2 \scriptsize{$\pm$0.3} \\ 
    \quad +MRC & 54.4  \scriptsize{$\pm$0.2} & 44.2 \scriptsize{$\pm$0.1} & 56.0 \scriptsize{$\pm$0.1} & 46.3 \scriptsize{$\pm$0.1} & 52.3 \scriptsize{$\pm$0.2} & 43.3 \scriptsize{$\pm$0.1}  & 56.0 \scriptsize{$\pm$0.1} & 46.2 \scriptsize{$\pm$0.2}  & 53.7 \scriptsize{$\pm$0.2} & 44.2 \scriptsize{$\pm$0.4} \\
    \quad +BT       & 57.5 \scriptsize{$\pm$0.7} & 54.0 \scriptsize{$\pm$0.9} & 61.2 \scriptsize{$\pm$0.3} & 57.9 \scriptsize{$\pm$0.1} & 57.3 \scriptsize{$\pm$0.5} & 54.0 \scriptsize{$\pm$0.6} & 61.6 \scriptsize{$\pm$0.8} & 58.1 \scriptsize{$\pm$0.7} & 57.6 \scriptsize{$\pm$0.5} & 54.3 \scriptsize{$\pm$0.5} \\
    
    \quad +TENT & 54.3 \scriptsize{$\pm$0.2} & 41.7 \scriptsize{$\pm$0.1} & 55.4 \scriptsize{$\pm$0.2} & 43.8 \scriptsize{$\pm$0.2} & 51.7 \scriptsize{$\pm$0.2} & 41.0 \scriptsize{$\pm$0.1} & 55.0 \scriptsize{$\pm$0.2} & 43.2 \scriptsize{$\pm$0.2} & 53.0 \scriptsize{$\pm$0.2} & 41.9 \scriptsize{$\pm$0.1} \\
    
    \quad +SAR & 54.5 \scriptsize{$\pm$0.5} & 41.5 \scriptsize{$\pm$0.4} & 54.9 \scriptsize{$\pm$0.3} & 43.1 \scriptsize{$\pm$0.2} & 51.0 \scriptsize{$\pm$0.4} & 40.3 \scriptsize{$\pm$0.6} & 55.3 \scriptsize{$\pm$0.5} & 43.0 \scriptsize{$\pm$0.6} & 52.9 \scriptsize{$\pm$0.2} & 41.4 \scriptsize{$\pm$0.4} \\
    
    \hline
    \multicolumn{11}{c}{\emph{Memory-Based Methods}} \\   
    
    TourHAMT        & 14.6 \scriptsize{$\pm$0.2} & 12.0 \scriptsize{$\pm$0.2} & 15.1 \scriptsize{$\pm$0.2} & 12.3 \scriptsize{$\pm$0.1} & 13.9 \scriptsize{$\pm$0.1} & 11.3 \scriptsize{$\pm$0.1} & 15.3 \scriptsize{$\pm$0.1} & 12.5 \scriptsize{$\pm$0.1} & 14.4 \scriptsize{$\pm$0.1} & 11.8 \scriptsize{$\pm$0.1} \\
    
    OVER-NAV        & 20.9 \scriptsize{$\pm$0.1} & 16.1 \scriptsize{$\pm$0.2} & 20.5 \scriptsize{$\pm$0.1} & 16.4 \scriptsize{$\pm$0.1} & 19.5 \scriptsize{$\pm$0.2} & 15.4 \scriptsize{$\pm$0.2} & 20.6 \scriptsize{$\pm$0.3} & 16.2 \scriptsize{$\pm$0.2} & 20.5 \scriptsize{$\pm$0.1} & 16.2 \scriptsize{$\pm$0.1}  \\
    GR-DUET   & 65.2 \scriptsize{$\pm$0.1} & 59.7 \scriptsize{$\pm$0.1} & 66.7 \scriptsize{$\pm$0.1} & 62.0 \scriptsize{$\pm$0.1} & 60.9 \scriptsize{$\pm$0.2} & \textbf{56.2} \scriptsize{$\pm$0.2} & 67.1 \scriptsize{$\pm$0.1} & 62.2 \scriptsize{$\pm$0.1} & 63.9 \scriptsize{$\pm$0.1} & 58.9 \scriptsize{$\pm$0.1} \\
        Ours   & \textbf{65.5} \scriptsize{$\pm$0.3} & \textbf{60.4} \scriptsize{$\pm$0.1} & \textbf{68.3} \scriptsize{$\pm$0.2} & \textbf{62.3} \scriptsize{$\pm$0.1} & \textbf{62.3} \scriptsize{$\pm$0.2} & 55.4 \scriptsize{$\pm$0.2} & \textbf{68.6} \scriptsize{$\pm$0.1} & \textbf{62.3} \scriptsize{$\pm$0.1} & \textbf{65.5} \scriptsize{$\pm$0.1} & \textbf{61.1} \scriptsize{$\pm$0.1} \\
        
  \bottomrule
  \end{tabular}
  }

\end{table*}

\section{Experiments}
\subsection{Experimental Setup}
\textbf{Datasets and Evaluation.} We follow the benchmark proposed by GR-DUET~\citep{GSA-VLN} and use the GSA-R2R dataset, which combines data from Habitat-Matterport3D (HM3D) and Matterport3D (MP3D). The dataset contains 150 evaluation scenes, including 75 in-distribution (ID) residential scenes and 75 out-of-distribution (OOD) non-residential scenes across 19 categories. 
In terms of language instructions, they include Basic instructions, as well as Scene instructions and User-style instructions generated by LLM simulating typical scene users or TV drama characters. For 600 paths in each scene, 7 instruction styles are generated, ultimately resulting in 90,000 path-instruction pairs.  
The splits are named using the format “Val/Test-R/N-Basic/Scene/User”, where “R” denotes residential and “N” represents non-residential scenes. We therefore adopt several evaluation metrics for navigation, including Navigation Error (NE, the distance between agent’s final location and the target location), Success Rate (SR), and SR penalized by Path Length (SPL), Trajectory Length (TL, the total navigation distance in meters), Normalized Dynamic Time Warping (nDTW, a measure of instruction fidelity by computing the similarity between the reference path and the predicted path). More details about GSA-R2R dataset can be seen in Supplementary Material.

\textbf{Implementation Details.} For the fast-reasoning component, we use the DUET~\citep{chen2022think} architecture. Image features are extracted using CLIP-ViT-B/16, and we employ 9 transformer layers in the text encoder. Other hyperparameters are set the same as in GR-DUET. What differs is that the visual observations obtained are converted into textual descriptions using llama3.2-vision~\citep{llama}. The entire slow-reasoning module is based on llama3.2-vision.

\subsection{Experimental Results}

\textbf{Environment Adaptation.}
We first tested these adaptation methods using basic instructions in different environments, with the results shown in Table \ref{tab:basic}. Compared to SOTA GR-DUET~\citep{GSA-VLN}, our method achieved the best performance on both the residential (R) and non-residential (N) datasets, with success rates (SR) improving by 1.5 \% and 2.2\%, respectively. This indicates that our fast reasoning module accumulates long-term memories related to scenarios, and the slow reasoning module refines them into generalized scenario rules and strategies, thereby empowering fast reasoning, helping the agent adapt to both in-distribution (ID) and out-of-distribution (OOD) environments.

\begin{table}
  \caption{Comparison of different adaptation methods in GSA-R2R with Scene instructions.}
  \label{tab:scene}
  \resizebox{\columnwidth}{!}{
  \begin{tabular}{l|ccccc}
    \toprule
    \multicolumn{1}{c|}{\multirow{2}{*}[-0.5ex]{\textbf{Methods}}} &  \multicolumn{5}{c}{\textbf{Test-N-Scene}} \\
    \cline{2-6}
    & TL & NE$\downarrow$  & SR$\uparrow$ & SPL$\uparrow$  & nDTW$\uparrow$ \\
    \hline
    \multicolumn{6}{c}{\emph{Baseline}} \\
    DUET            & 14.9 & 6.4 & 39.6 & 30.1 & 40.9  \\
    \hline
    \multicolumn{6}{c}{\emph{Optimization-Based Methods}} \\   
    \quad +MLM      & 14.3 \scriptsize{$\pm$0.1} & 6.5 \scriptsize{$\pm$0.1} & 39.8 \scriptsize{$\pm$0.1} & 30.5 \scriptsize{$\pm$0.1} & 41.1 \scriptsize{$\pm$0.1} \\
    
    \quad +MRC      & 14.9 \scriptsize{$\pm$0.1} & 6.4 \scriptsize{$\pm$0.1} & 39.7 \scriptsize{$\pm$0.1} & 30.2 \scriptsize{$\pm$0.1} & 40.9 \scriptsize{$\pm$0.1} \\

    \quad +BT       & 8.4 \scriptsize{$\pm$0.0} & 6.3 \scriptsize{$\pm$0.2} & 41.2 \scriptsize{$\pm$1.5} & 38.2 \scriptsize{$\pm$1.2} & 51.3 \scriptsize{$\pm$1.2} \\

    \quad +TENT     & 16.4 \scriptsize{$\pm$0.1} & 6.3 \scriptsize{$\pm$0.1} & 40.6 \scriptsize{$\pm$0.2} & 28.9 \scriptsize{$\pm$0.2} & 38.9 \scriptsize{$\pm$0.2} \\
    
    \quad +SAR & 16.3 \scriptsize{$\pm$0.5} & 6.0 \scriptsize{$\pm$0.2} & 41.4 \scriptsize{$\pm$0.6} & 29.1 \scriptsize{$\pm$0.3} & 39.0 \scriptsize{$\pm$0.3} \\
    
    \hline
    \multicolumn{6}{c}{\emph{Memory-Based Methods}} \\   
    TourHAMT        & 7.3 \scriptsize{$\pm$0.1} & 8.1 \scriptsize{$\pm$0.1} & 9.7 \scriptsize{$\pm$0.1} & 8.0 \scriptsize{$\pm$0.1} & 32.3 \scriptsize{$\pm$0.1} \\
    OVER-NAV        & 11.8 \scriptsize{$\pm$0.1} & 7.6 \scriptsize{$\pm$0.2} & 16.7 \scriptsize{$\pm$0.4} & 12.6 \scriptsize{$\pm$0.2} & 34.6 \scriptsize{$\pm$0.3} \\
    GR-DUET  & 10.1 \scriptsize{$\pm$0.0} & 5.5 \scriptsize{$\pm$0.0} & 48.1 \scriptsize{$\pm$0.1} & 42.8 \scriptsize{$\pm$0.1} & 53.7 \scriptsize{$\pm$0.1} \\
Ours  & 8.9 \scriptsize{$\pm$0.2} & \textbf{5.1} \scriptsize{$\pm$0.0} & \textbf{50.7} \scriptsize{$\pm$0.1} & \textbf{46.6} \scriptsize{$\pm$0.1} & \textbf{57.8} \scriptsize{$\pm$0.3} \\
    
    \bottomrule
  \end{tabular}

  }
\end{table}

\begin{table}
  \centering
  \caption{Analysis of ablation experiments on each module.}
  \label{tab:ablation-1}
    \resizebox{\columnwidth}{!}{ 
  \begin{tabular}{c|c|cc|cc|cc}
    \toprule
     \multicolumn{1}{c|}{\multirow{2}{*}{FSR}} & \multicolumn{1}{c|}{\multirow{2}{*}{ISC}} & \multicolumn{2}{c|}{\textbf{Test-R-Basic}} & \multicolumn{2}{c|}{\textbf{Test-N-Basic}} & \multicolumn{2}{c}{\textbf{Test-N-Scene}}  \\
    \cline{3-8}
    & & SR$\uparrow$ & SPL$\uparrow$  & SR$\uparrow$ & SPL$\uparrow$  & SR$\uparrow$ & SPL$\uparrow$ \\
    \midrule
    $\times$ & $\times$           & 64.0 \scriptsize{$\pm$0.1} & 58.0 \scriptsize{$\pm$0.2} & 53.7 \scriptsize{$\pm$0.2} & 47.5 \scriptsize{$\pm$0.1} & 42.4 \scriptsize{$\pm$0.1} & 42.8 \scriptsize{$\pm$0.2} \\
    $\times$ & $\checkmark$       & 64.0 \scriptsize{$\pm$0.1} & 58.0 \scriptsize{$\pm$0.2} & 53.7 \scriptsize{$\pm$0.2} & 47.5 \scriptsize{$\pm$0.1} & 46.1 \scriptsize{$\pm$0.4} & 44.8 \scriptsize{$\pm$0.0} \\
    $\checkmark$ & $\times$       & 69.1 \scriptsize{$\pm$0.1} & 63.9 \scriptsize{$\pm$0.2} & 58.4 \scriptsize{$\pm$0.1} & 52.9 \scriptsize{$\pm$0.1} & 47.9 \scriptsize{$\pm$0.2} & 45.0 \scriptsize{$\pm$0.2} \\
    $\checkmark$ & $\checkmark$   & \textbf{69.1} \scriptsize{$\pm$0.1} & \textbf{63.9} \scriptsize{$\pm$0.2} & \textbf{58.4} \scriptsize{$\pm$0.1} & \textbf{52.9} \scriptsize{$\pm$0.1} & \textbf{50.4} \scriptsize{$\pm$0.1} & \textbf{46.4} \scriptsize{$\pm$0.1} \\
  \bottomrule
  \end{tabular}
  }
\end{table}

\begin{table}
  \caption{Analysis of the impact of $K$.}
  \label{tab:ablation-2}
  \vspace{-5pt}
    \resizebox{\columnwidth}{!}{ 
  \begin{tabular}{c|cc|cc|cc}
    \toprule
     \multicolumn{1}{c|}{\multirow{2}{*}{$K$}} & \multicolumn{2}{c|}{\textbf{Test-R-Basic}} & \multicolumn{2}{c|}{\textbf{Test-N-Basic}} & \multicolumn{2}{c}{\textbf{Test-N-Scene}}  \\
    \cline{2-7}
    & SR$\uparrow$ & SPL$\uparrow$  & SR$\uparrow$ & SPL$\uparrow$  & SR$\uparrow$ & SPL$\uparrow$ \\
    \midrule
    20           & 61.3 \scriptsize{$\pm$0.2} & 54.7 \scriptsize{$\pm$0.1} & 47.5 \scriptsize{$\pm$0.2} & 44.0 \scriptsize{$\pm$0.0} & 43.6 \scriptsize{$\pm$0.2} & 40.2 \scriptsize{$\pm$0.2} \\
    50        & \textbf{65.4} \scriptsize{$\pm$0.1} & \textbf{64.9} \scriptsize{$\pm$0.1}& 54.9 \scriptsize{$\pm$0.0} & 47.4 \scriptsize{$\pm$0.1} & 47.5 \scriptsize{$\pm$0.1} & 44.0 \scriptsize{$\pm$0.3}\\
    100           & 62.0 \scriptsize{$\pm$0.2} & 57.6 \scriptsize{$\pm$0.0}  & \textbf{60.2} \scriptsize{$\pm$0.2} & \textbf{52.0} \scriptsize{$\pm$0.1} & \textbf{48.5} \scriptsize{$\pm$0.1} & \textbf{45.6} \scriptsize{$\pm$0.1}  \\
    200      & 63.1 \scriptsize{$\pm$0.1} & 64.0 \scriptsize{$\pm$0.0} & 58.0 \scriptsize{$\pm$0.1} & 51.6 \scriptsize{$\pm$0.0} & 46.3 \scriptsize{$\pm$0.3} & 44.7 \scriptsize{$\pm$0.0} \\
  \bottomrule
  \end{tabular}
  }
  \vspace{-5pt}
\end{table}

\textbf{Instruction Adaptation.}
We evaluated these methods under different instruction styles. Table \ref{tab:user} presents the results for the model under five user instructions roles, while Table \ref{tab:scene} shows their performance with scene instructions. First, DUET's~\citep{chen2022think} performance data indicates that different expression styles introduce varying levels of difficulty in instruction interpretation for VLN models. Second, our method outperforms GR-DUET~\citep{GSA-VLN} in both scene and user-style instructions. The key innovation lies in addressing GR-DUET's limitation in adapting to instruction styles. We achieve this through an LLM-based instruction style conversion mechanism (using prompt engineering to transform scene/user styles into the model's familiar Basic style while retaining core semantics). Meanwhile, we incorporated a dynamic feedback loop where quick reasoning accumulates “scene-instruction-action” memories, while slow reasoning refines structured knowledge, which then feeds back into quick decision-making. As a result, our approach achieves better navigation success rates, path accuracy, and other metrics under both instruction styles.

\subsubsection{Ablation Study}

\textbf{Component Analysis.} The main modules we propose consist of two parts: the Fast-Slow Reasoning (FSR) framework and Instruction Style Conversion (ISC). We conducted ablation experiments on them, as shown in Table \ref{tab:ablation-1}. First, compared with the first row, when ISC is added in the second row, it can be seen that Test-R-Basic and Test-N-Basic remain unchanged, while the performance of Test-N-Scene has improved. This is because the instruction style conversion model works on Scene-style instructions. In the third row, when only FST is added, compared with the first two rows, the performance of each column has improved, which indicates that our fast-slow thinking framework can improve the performance of all types of instructions, verifying its effectiveness. In the fourth row, when FST and ISC work together, the performance of Test-N-Scene reaches the best, verifying the collaborative effectiveness of FST and ISC.

\begin{figure}  
\centering    
\includegraphics[width=\columnwidth]{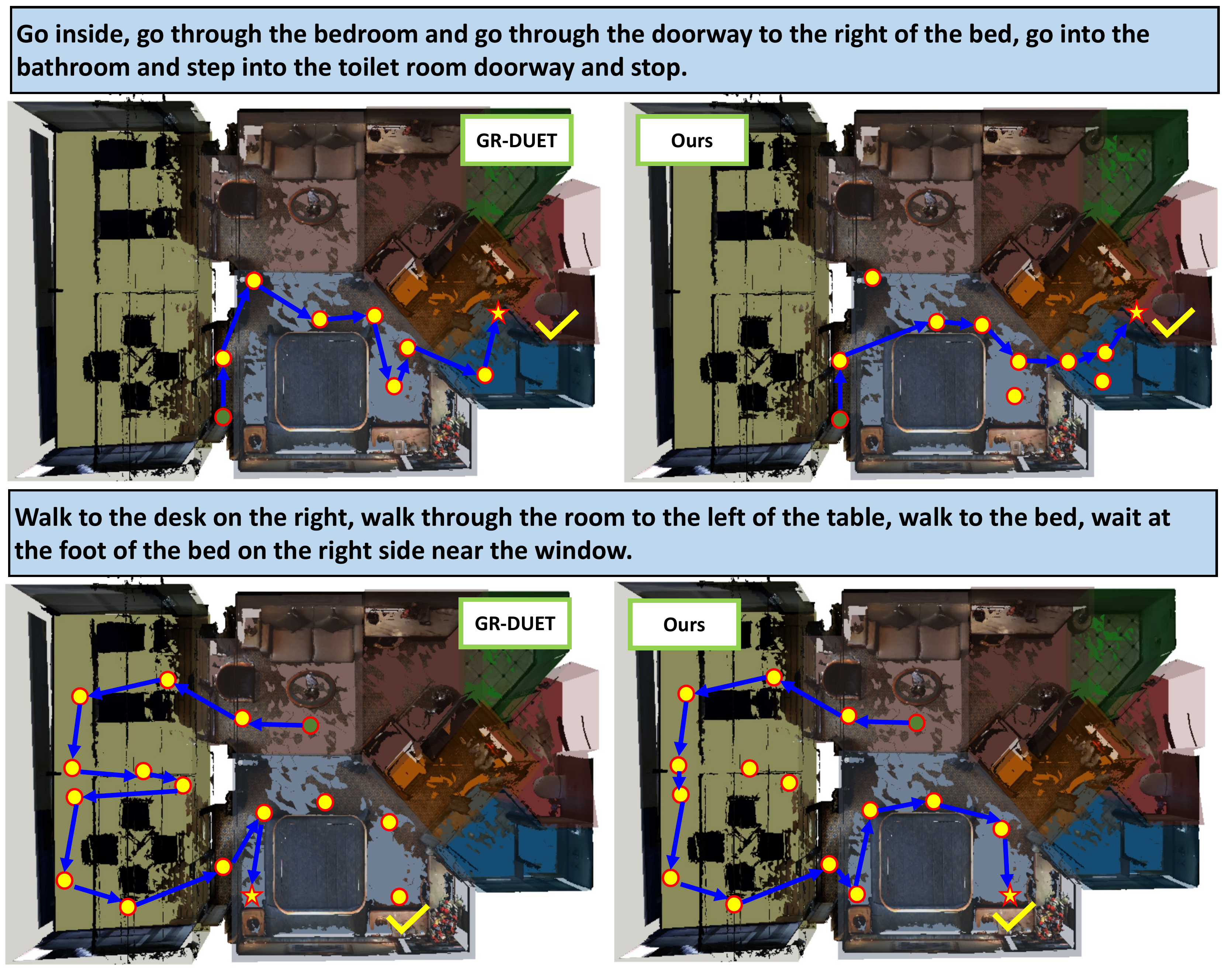}    
\caption{Predicted trajectories of GR-DUET (left) and our method (right). A check mark ($\checkmark$) indicates the destination; A five-pointed star ($\star$) marks the final position reached by the agent.}   
\vspace{-16pt}
\label{fig:sota-vis}
\end{figure}

We visualized the predicted trajectories of GR-DUET~\citep{GSA-VLN} and our method (see Fig. \ref{fig:sota-vis}): In the scenarios shown in the first row, both methods reached the correct destination, but GR-DUET exhibited inefficient, excessively long trajectories with unnecessary detours due to its lack of accumulated scene experience. In the second row, GR-DUET incorrectly stopped to the left of the bed-mistaking a bedside lamp for a window and failed to reach the target, with its trajectory also being overly redundant. By contrast, our method efficiently arrived at the window to the right of the bed. This advantage stems from the scene spatial rules (e.g., “window on the right wall of the bed”) and the experience loop in our slow-thinking module. It helps avoid detours by using the experience library and accurately identifies targets through a combination of real-time observations and scene priors.


\textbf{Experience Library Capacity $K$.} 
This experiment verifies the capacity saturation effect of experience library capacity \(K\): whether insufficient storage limits generalization when \(K\) is too small, and whether redundant experience causes surging computational overhead and low-quality interference when \(K\) is too large, ultimately determining the optimal \(K\) range.
From Table \ref{tab:ablation-2}, performance across all scenarios is lowest when \(K=20\), as the experience library fails to store key rules for OOD scenarios, restricting generalization. Test-R-Basic performs best at \(K=50\), since core experiences for basic instructions in residential scenes are sufficiently stored, and further increasing \(K\) adds redundancy. Test-N-Basic and Test-N-Scene achieve optimal performance at \(K=100\), as OOD non-residential scenes require more generalized experiences. Performance declines at \(K=200\), possibly due to redundant experiences interfering with attention fusion.  
In summary, \(K<50\) leads to insufficient experience, while \(K>100\) causes redundancy. The optimal \(K\) ranges from 50 to 100: we can use 100 for complex OOD scenarios and 50 for low-to-medium complexity scenarios.

\subsection{Case Study: Before and After Slow Reasoning}
To illustrate the qualitative advantages of the slow4fast architecture, we compare and analyze the situations of using fast reasoning alone and the interaction between slow and fast reasoning. The instruction is: “\textit{Leave the kitchen and take a right into the hallway. In the hall take the right into the den, then a left into the dining room. In the dining room stop next to the door near the vent in the floor.}” 
The key spatial viewpoints are as follows: $V_1$: Inside the kitchen (starting point); $V_2$: Kitchen exit, connecting to the hallway;
$V_3$: Hallway, with branches leading to $V_4$ (den entrance); $V_4$: Den entrance, leading to $V_5$ (left door to dining room); $V_6$: Bathroom, with a door leading to $V_7$ (dining room); $V_7$: Dining room, near the door and floor vent (destination).




\textbf{Initial Challenges.} The hallway has multiple branches, making it easy to go to the den and walk around in circles without prior experience; The visual feature “door near the vent” in the dining room is not prominent (the vent is small and partially obscured by a rug), making it easy to miss the target location during the first navigation attempt. At this stage, the experience library is empty. The fast-reasoning module relies solely on real-time vision and instructions, causing navigation errors.

\textbf{Navigation Trajectory (Initial Attempt).} $V_1$ (Inside Kitchen) $\rightarrow$ Leaves kitchen to $V_2$ (Kitchen Exit), turns right into the hallway ($V_3$) $\rightarrow$ Due to the dim lighting and multiple branches in the hallway, mistakenly selects the middle branch (not the correct right turn to the den), proceeds to the end of the hallway, and then turns back $\rightarrow$ Finds the correct right turn and enters the den ($V_4$) $\rightarrow$ After entering the den, fails to identify the left door to the dining room (partially blocked by a bookshelf) and wanders in a 1.2m circle $\rightarrow$ Finds the door and enters the bathroom ($V_6$) $\rightarrow$ Mistakenly identifies a common cabinet in the bathroom as the "door near the vent" and stops (fails to reach $V_7$).
The total time consumed for navigation was 15 seconds, the navigation error reached 1.5 meters. The core issue is the lack of prior knowledge of spatial features, such as the right turn to the den and the vent near the dining room door, leading to unnecessary detours and misidentifications.

\textbf{Experience Distillation (Post-Navigation Reflection).} The failure log from the first navigation is stored in the history repository, and the slow-reasoning module initiates a reflection process. The log is input into an LLM to generate experience $E_1$, which is stored in the experience library. 

$S_t$ (\textit{Scene Type}): residential-kitchen to dining room transition area.  

$C_s$ (\textit{Spatial Context}): Correct path in the hallway marked by a blue painting; the vent in the dining room is square and fully exposed, with a wooden door beside it.

$R_s$ (\textit{Spatial Rules}): In the hallway, look for the blue painting on the second right branch; in the dining room, find the vent before the wooden door.

$T_n$ (\textit{Navigation Strategy}): Upon reaching $V_3$ (Hallway), scan for the blue painting; in the dining room, locate the vent and the wooden door next to it.

After 4 iterations, the Experience Library has accumulated 6 similar experiences (e.g., “the effect of changing light in the hallway on branch identification,” “reinforcing features of the left door in the den”). During the 5th navigation, the fast-reasoning module is empowered by this experience.

\textbf{Navigation Trajectory (Post-Experience).} $V_1$ (Inside Kitchen) $\rightarrow$$V_2$$\rightarrow$$V_3$$\rightarrow$ Guided by the “blue painting” experience, quickly finds the second branch on the right (the correct path to the den) $\rightarrow$ Enters the den ($V_4$) and, based on the experience “the left door is next to a round desk," directly finds the left door to the dining room ($V_5$) $\rightarrow$ Enters the bathroom ($V_6$) and finds the right door leading to the dining room $\rightarrow$ Enters the dining room, locates the fully exposed square vent, and stops next to the wooden door beside it ($V_7$)  (target).
The total time taken was 8 seconds, representing a 46.7\% reduction; the navigation error was minimized to 0.3 meters, an 80\% reduction from the initial attempt; 
The improvements stem from the agent’s experience, enabling it to identify key features, leading to a more efficient path.

\begin{figure}
  \includegraphics[width=\linewidth]{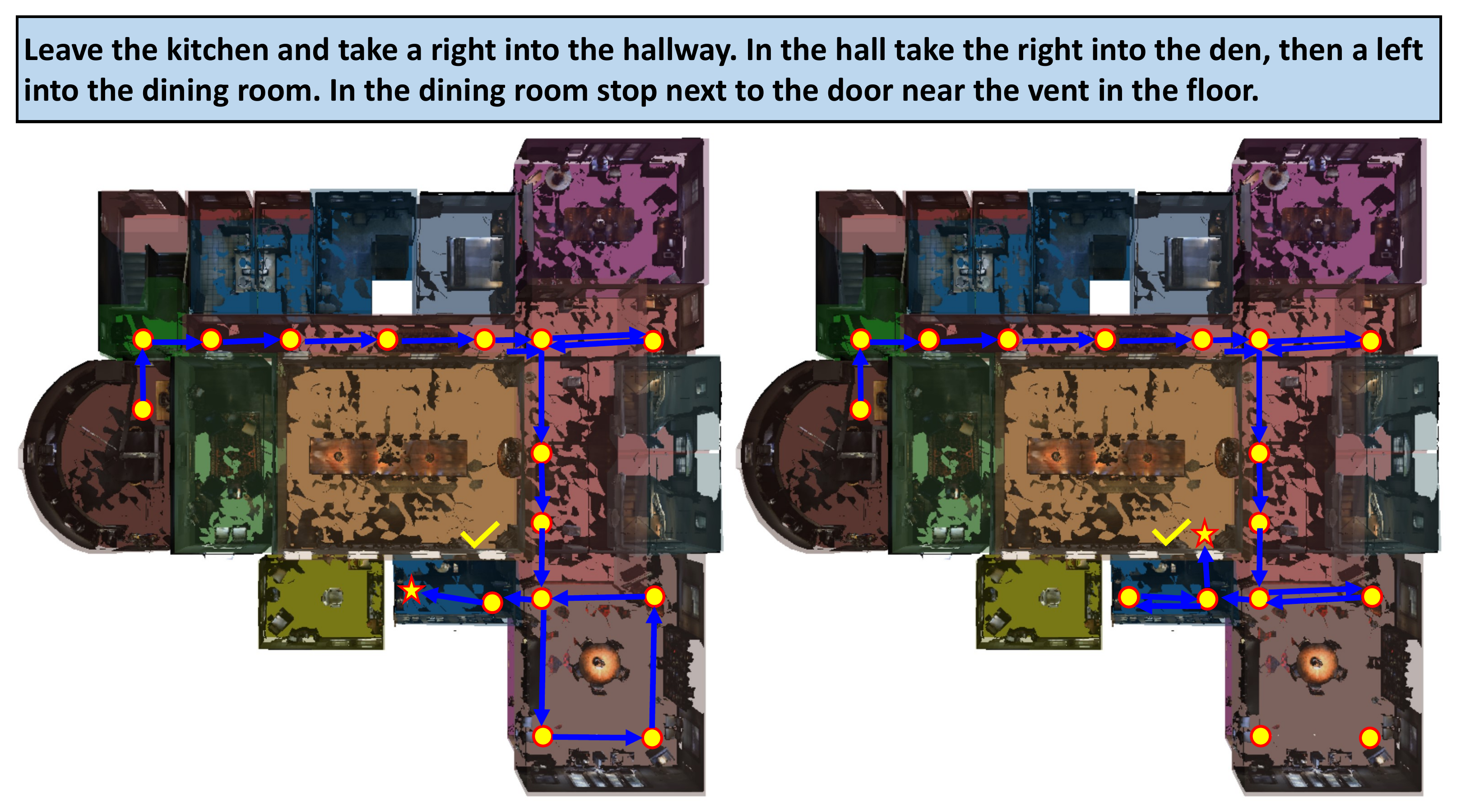}
\centering
  \caption{\textbf{Case Study.} The left side shows the execution trajectory of the agent with \textbf {fast reasoning only}, while the right side displays the agent's trajectory \textbf {after slow reasoning}. A check mark ($\checkmark$) indicates the destination (next to the door near the floor vent); A five-pointed star ($\star$) marks the final position reached by the agent.
}
    \label{casestudy}
    \vspace{-11pt}
\end{figure}
\section{Conclusion}
In this paper, we focus on the GSA-VLN task, aiming to learn generalized navigation ability by introducing diverse environments and inconsistent intructions.
Inspired by the fast-slow cognition systems, we propose the slow4fast-VLN, establishing a dynamic interactive fast-slow reasoning framework.
First, we receive input data, and the fast-reasoning module generating immediate navigation actions and storing memories. The slow-reasoning module analyzes these memories, extracts generalized experiences through deep reflection, and structurally stores them to optimize the fast module. This enabling the system to adapt and execute navigation efficiently in unseen scenarios.

\textbf{Acknowledgment.} This work is supported by the National Nature Science Foundation of China (Nos. 62376186, 62472333).

{
    \small
    \bibliographystyle{ieeenat_fullname}
    \bibliography{main}
}


\end{document}